\documentclass[english]{article}
\usepackage[T1]{fontenc}
\usepackage[latin9]{inputenc}
\usepackage{wrapfig}
\usepackage{booktabs}
\usepackage{amsmath}
\usepackage{amssymb}
\usepackage{graphicx}
\usepackage[numbers]{natbib}

\makeatletter

\providecommand{\tabularnewline}{\\}

\newenvironment{lyxcode}
{\par\begin{list}{}{
\setlength{\rightmargin}{\leftmargin}
\setlength{\listparindent}{0pt}% needed for AMS classes
\raggedright
\setlength{\itemsep}{0pt}
\setlength{\parsep}{0pt}
\normalfont\ttfamily}%
 \item[]}
{\end{list}}

\usepackage[final]{nips_2016}

\@ifundefined{showcaptionsetup}{}{%
 \PassOptionsToPackage{caption=false}{subfig}}
\usepackage{subfig}
\makeatother

\usepackage{babel}
\begin{document}
\begin{lyxcode}

\end{lyxcode}

\title{BEGAN: Boundary Equilibrium Generative Adversarial Networks}

\author{David Berthelot, Thomas Schumm, Luke Metz\\ 
Google\\
\texttt{\{dberth,fwiffo,lmetz\}@google.com}}
\maketitle
\begin{abstract}
We propose a new equilibrium enforcing method paired with a loss derived
from the Wasserstein distance for training auto-encoder based Generative
Adversarial Networks. This method balances the generator and discriminator
during training. Additionally, it provides a new approximate convergence
measure, fast and stable training and high visual quality. We also
derive a way of controlling the trade-off between image diversity
and visual quality. We focus on the image generation task, setting
a new milestone in visual quality, even at higher resolutions. This
is achieved while using a relatively simple model architecture and
a standard training procedure. 
\end{abstract}

\section{Introduction}

Generative Adversarial Networks \cite{goodfellow2014generative_gan}(GANs)
are a class of methods for learning a data distribution $p_{model}(x)$
and realizing a model to sample from it. GANs are architectured around
two functions: the generator $G(z)$, which maps a sample $z$ from
a random uniform distribution to the data distribution, and the discriminator
$D(x)$ which determines if a sample $x$ belongs to the data distribution.
The generator and discriminator are typically learned jointly by alternating
the training of $D$ and $G$, based on game theory principles. 

GANs can generate very convincing images, sharper than ones produced
by auto-encoders using pixel-wise losses. However, GANs still face
many unsolved difficulties: in general they are notoriously difficult
to train, even with many tricks applied \cite{radford2015unsupervised_dcgan,salimans2016improved}.
Correct hyper-parameter selection is critical. Controlling the image
diversity of the generated samples is difficult. Balancing the convergence
of the discriminator and of the generator is a challenge: frequently
the discriminator wins too easily at the beginning of training \cite{goodfellow2016nips_gantut}.
GANs easily suffer from modal collapse, a failure mode in which just
one image is learned \cite{dumoulin2016adversarially}. Heuristic
regularizers such as batch discrimination \cite{salimans2016improved}
and the repelling regularizer \cite{zhao2016energy_ebgan} have been
proposed to alleviate this problem with varying degrees of success. 

In this paper, we make the following contributions:
\begin{itemize}
\item A GAN with a simple yet robust architecture, standard training procedure
with fast and stable convergence.
\item An equilibrium concept that balances the power of the discriminator
against the generator. 
\item A new way to control the trade-off between image diversity and visual
quality.
\item An approximate measure of convergence. To our knowledge the only other
published measure is from Wasserstein GAN \cite{arjovsky2017wasserstein}
(WGAN), which will be discussed in the next section.
\end{itemize}

\section{Related work}

Deep Convolutional GANs \cite{radford2015unsupervised_dcgan}(DCGANs)
first introduced a convolutional architecture which led to improved
visual quality. More recently, Energy Based GANs \cite{zhao2016energy_ebgan}(EBGANs)
were proposed as a class of GANs that aims to model the discriminator
$D(x)$ as an energy function. This variant converges more stably
and is both easy to train and robust to hyper-parameter variations.
The authors attribute some of these benefits to the larger number
of targets in the discriminator. EBGAN likewise implements its discriminator
as an auto-encoder with a per-pixel error.

While earlier GAN variants lacked a measure of convergence, Wasserstein
GANs \cite{arjovsky2017wasserstein} (WGANs) recently introduced a
loss that also acts as a measure of convergence. In their implementation
it comes at the expense of slow training, but with the benefit of
stability and better mode coverage.

\section{Proposed method}

We use an auto-encoder as a discriminator as was first proposed in
EBGAN \cite{zhao2016energy_ebgan}. While typical GANs try to match
data distributions directly, our method aims to match auto-encoder
loss distributions using a loss derived from the Wasserstein distance.
This is done using a typical GAN objective with the addition of an
equilibrium term to balance the discriminator and the generator. Our
method has an easier training procedure and uses a simpler neural
network architecture compared to typical GAN techniques.

\subsection{Wasserstein distance lower bound for auto-encoders}

We wish to study the effect of matching the distribution of the errors
instead of matching the distribution of the samples directly. We first
introduce the auto-encoder loss, then we compute a lower bound to
the Wasserstein distance between the auto-encoder loss distributions
of real and generated samples. 

We first introduce $\mathcal{L}:\mathbb{R}^{N_{x}}\mapsto\mathbb{R}^{+}$the
loss for training a pixel-wise autoencoder as:

\[
\mathcal{L}(v)=|v-D(v)|^{\eta}\textrm{ where }\begin{cases}
D:\mathbb{R}^{N_{x}}\mapsto\mathbb{R}^{N_{x}} & \textrm{is the autoencoder function.}\\
\eta\in\left\{ 1,2\right\}  & \textrm{is the target norm}.\\
v\in\mathbb{R}^{N_{x}} & \textrm{is a sample of dimension \ensuremath{N_{x}}.}
\end{cases}
\]

Let $\mu_{1,2}$ be two distributions of auto-encoder losses, let
$\Gamma(\mu_{1},\mu_{2})$ be the set all of couplings of $\mu_{1}$
and $\mu_{2}$, and let $m_{1,2}\in\mathbb{R}$ be their respective
means. The Wasserstein distance can be expressed as:

\[
W_{1}(\mu_{1},\mu_{2})=\inf_{\gamma\in\Gamma(\mu_{1},\mu_{2})}\mathbb{E}_{(x_{1},x_{2})\sim\gamma}[|x_{1}-x_{2}|]
\]
 Using Jensen's inequality, we can derive a lower bound to $W_{1}(\mu_{1},\mu_{2})$:

\begin{equation}
\inf\,\mathbb{E}[|x_{1}-x_{2}|]\geqslant\inf\,|\mathbb{E}[x_{1}-x_{2}]|=|m_{1}-m_{2}|\label{eq:lowerbound}
\end{equation}

It is important to note that we are aiming to optimize a lower bound
of the Wasserstein distance between auto-encoder loss distributions,
not between sample distributions.

\subsection{GAN objective}

We design the discriminator to maximize equation \ref{eq:lowerbound}
between auto-encoder losses. Let $\mu_{1}$ be the distribution of
the loss $\mathcal{L}(x)$, where $x$ are real samples. Let $\mu_{2}$
be the distribution of the loss $\mathcal{L}(G(z))$, where $G:\mathbb{R}^{N_{z}}\mapsto\mathbb{R}^{N_{x}}$
is the generator function and $z\in[-1,1]^{N_{z}}$ are uniform random
samples of dimension $N_{z}$.

Since $m_{1},m_{2}\in\mathbb{R}^{+}$ there are only two possible
solutions to maximizing $|m_{1}-m_{2}|$:

\[
(a)\begin{cases}
W_{1}(\mu_{1},\mu_{2})\geqslant m_{1}-m_{2}\\
m_{1}\rightarrow\infty\\
m_{2}\rightarrow0
\end{cases}\textrm{ or }(b)\begin{cases}
W_{1}(\mu_{1},\mu_{2})\geqslant m_{2}-m_{1}\\
m_{1}\rightarrow0\\
m_{2}\rightarrow\infty
\end{cases}
\]

We select solution (b) for our objective since minimizing $m_{1}$
leads naturally to auto-encoding the real images. Given the discriminator
and generator parameters $\theta_{D}$ and $\theta_{G}$, each updated
by minimizing the losses $\mathcal{L}_{D}$ and $\mathcal{L}_{G}$,
we express the problem as the GAN objective, where $z_{D}$ and $z_{G}$
are samples from $z$:

\begin{equation}
\begin{cases}
\mathcal{L}_{D}=\mathcal{L}(x;\theta_{D})-\mathcal{L}(G(z_{D};\theta_{G});\theta_{D}) & \textrm{for }\theta_{D}\\
\mathcal{L}_{G}=-\mathcal{L}_{D} & \textrm{for }\theta_{G}
\end{cases}\label{eq:wbgan}
\end{equation}

Note that in the following we use an abbreviated notation: $G(\cdot)=G(\cdot,\theta_{G})$
and $\mathcal{L}(\cdot)=\mathcal{L}(\cdot;\theta_{D})$.

This equation, while similar to the one from WGAN \cite{arjovsky2017wasserstein},
has two important differences: First we match distributions between
losses, not between samples. And second, we do not explicitly require
the discriminator to be K-Lipschitz since we are not using the Kantorovich
and Rubinstein duality theorem \cite{villani2008optimal}.

For function approximations, in our case deep neural networks, we
must also consider the representational capacities of each function
$G$ and $D$. This is determined both by the model implementing the
function and the number of parameters. It is typically the case that
$G$ and $D$ are not well balanced and the discriminator $D$ wins
easily. To account for this situation we introduce an equilibrium
concept.

\subsection{Equilibrium}

In practice it is crucial to maintain a balance between the generator
and discriminator losses; we consider them to be at equilibrium when:

\begin{equation}
\mathbb{E}\left[\mathcal{L}(x)\right]=\mathbb{E}\left[\mathcal{L}(G(z))\right]\label{eq:equilibrium}
\end{equation}

If we generate samples that cannot be distinguished by the discriminator
from real ones, the distribution of their errors should be the same,
including their expected error. This concept allows us to balance
the effort allocated to the generator and discriminator so that neither
wins over the other.

We can relax the equilibrium with the introduction of a new hyper-parameter
$\gamma\in[0,1]$ defined as

\begin{equation}
\gamma=\frac{\mathbb{E}\left[\mathcal{L}(G(z))\right]}{\mathbb{E}\left[\mathcal{L}(x)\right]}\label{eq:gamma}
\end{equation}

In our model, the discriminator has two competing goals: auto-encode
real images and discriminate real from generated images. The $\gamma$
term lets us balance these two goals. Lower values of $\gamma$ lead
to lower image diversity because the discriminator focuses more heavily
on auto-encoding real images. We will refer to $\gamma$ as the diversity
ratio. There is a natural boundary for which images are sharp and
have details.

\subsection{Boundary Equilibrium GAN}

The BEGAN objective is:

\[
\begin{cases}
\mathcal{L}_{D}=\mathcal{L}(x)-k_{t}.\mathcal{L}(G(z_{D})) & \textrm{for }\theta_{D}\\
\mathcal{L}_{G}=\mathcal{L}(G(z_{G})) & \textrm{for }\theta_{G}\\
k_{t+1}=k_{t}+\lambda_{k}(\gamma\mathcal{L}(x)-\mathcal{L}(G(z_{G}))) & \textrm{for each training step }t
\end{cases}
\]

We use Proportional Control Theory to maintain the equilibrium $\mathbb{E}\left[\mathcal{L}(G(z))\right]=\gamma\mathbb{E}\left[\mathcal{L}(x)\right]$.
This is implemented using a variable $k_{t}\in\left[0,1\right]$ to
control how much emphasis is put on $\mathcal{L}(G(z_{D}))$ during
gradient descent. We initialize $k_{0}=0$. $\lambda_{k}$ is the
proportional gain for $k$; in machine learning terms, it is the learning
rate for $k$. We used 0.001 in our experiments. In essence, this
can be thought of as a form of closed-loop feedback control in which
$k_{t}$ is adjusted at each step to maintain equation \ref{eq:gamma}.

In early training stages, $G$ tends to generate easy-to-reconstruct
data for the auto-encoder since generated data is close to 0 and the
real data distribution has not been learned accurately yet. This yields
to $\mathcal{L}(x)>\mathcal{L}(G(z))$ early on and this is maintained
for the whole training process by the equilibrium constraint. 

The introductions of the approximation in equation \ref{eq:lowerbound}
and $\gamma$ in equation \ref{eq:gamma} have an impact on our modeling
of the Wasserstein distance. Consequently, examination of samples
generated from various $\gamma$ values is of primary interest as
will be shown in the results section. 

In contrast to traditional GANs which require alternating training
$D$ and $G$, or pretraining $D$, our proposed method BEGAN requires
neither to train stably. Adam \cite{kingma2014adam} was used during
training with the default hyper-parameters. $\theta_{D}$ and $\theta_{G}$
are updated independently based on their respective losses with separate
Adam optimizers. We typically used a batch size of $n=16$.

\subsubsection{Convergence measure}

Determining the convergence of GANs is generally a difficult task
since the original formulation is defined as a zero-sum game. As a
consequence, one loss goes up when the other goes down. The number
of epochs or visual inspection are typically the only practical ways
to get a sense of how training has progressed.

We derive a global measure of convergence by using the equilibrium
concept: we can frame the convergence process as finding the closest
reconstruction $\mathcal{L}(x)$ with the lowest absolute value of
the instantaneous process error for the proportion control algorithm
$|\gamma\mathcal{L}(x)-\mathcal{L}(G(z_{G}))|$. This measure is formulated
as the sum of these two terms:

\[
\mathcal{M}{}_{global}=\mathcal{L}(x)+|\gamma\mathcal{L}(x)-\mathcal{L}(G(z_{G}))|
\]

This measure can be used to determine when the network has reached
its final state or if the model has collapsed.

\subsection{Model architecture}

The discriminator $D:\mathbb{R}^{N_{x}}\mapsto\mathbb{R}^{N_{x}}$
is a convolutional deep neural network architectured as an auto-encoder.
$N_{x}=H\times W\times C$ is shorthand for the dimensions of $x$
where $H,W,C$ are the height, width and colors. We use an auto-encoder
with both a deep encoder and decoder. The intent is to be as simple
as possible to avoid typical GAN tricks. 

\begin{figure}

\subfloat[Generator/Decoder]{\noindent \begin{centering}
\includegraphics[bb=0bp 0bp 661bp 425bp,clip,height=5cm]{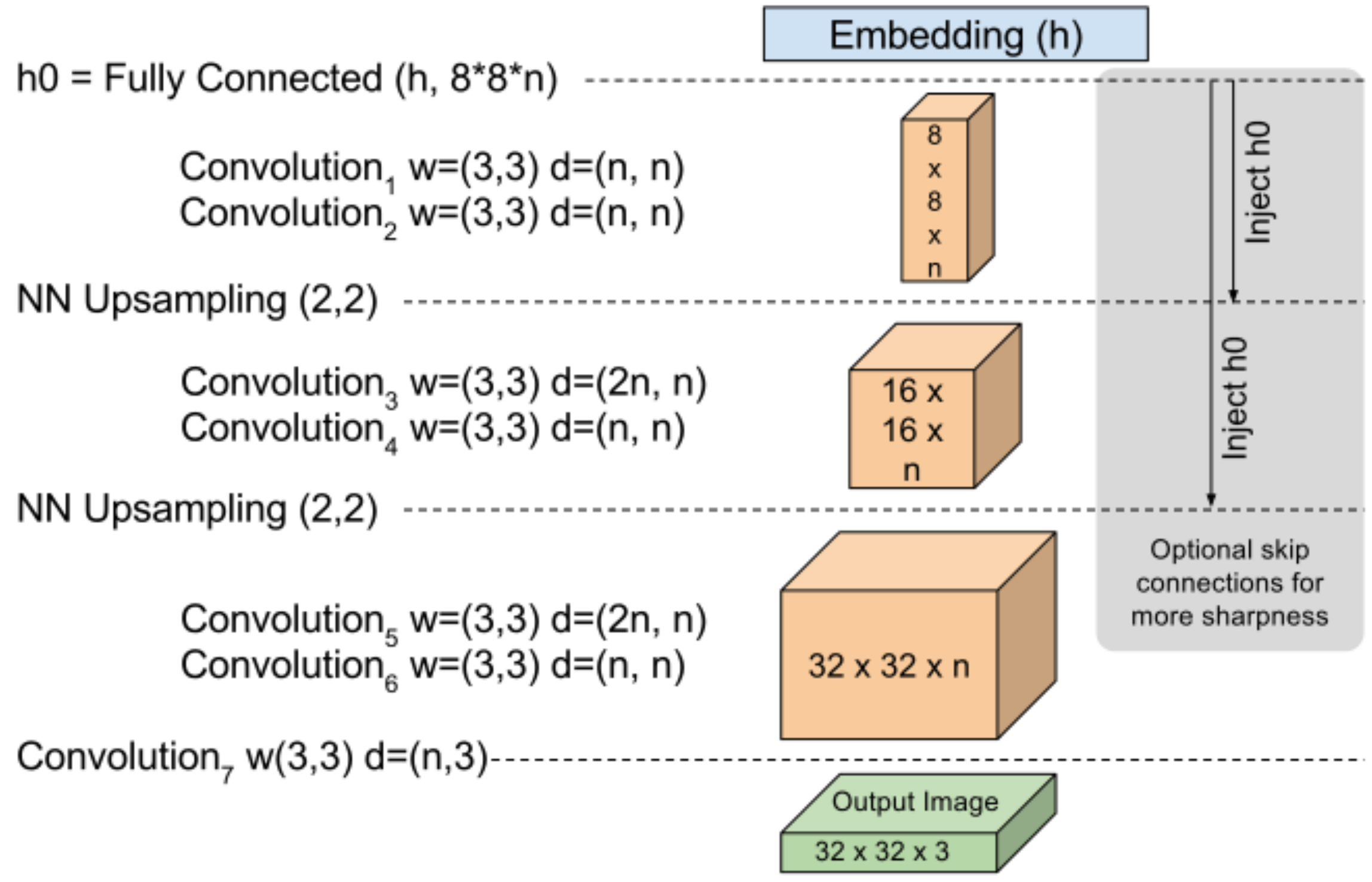}
\par\end{centering}

}\subfloat[Encoder]{\begin{centering}
\includegraphics[clip,height=5cm]{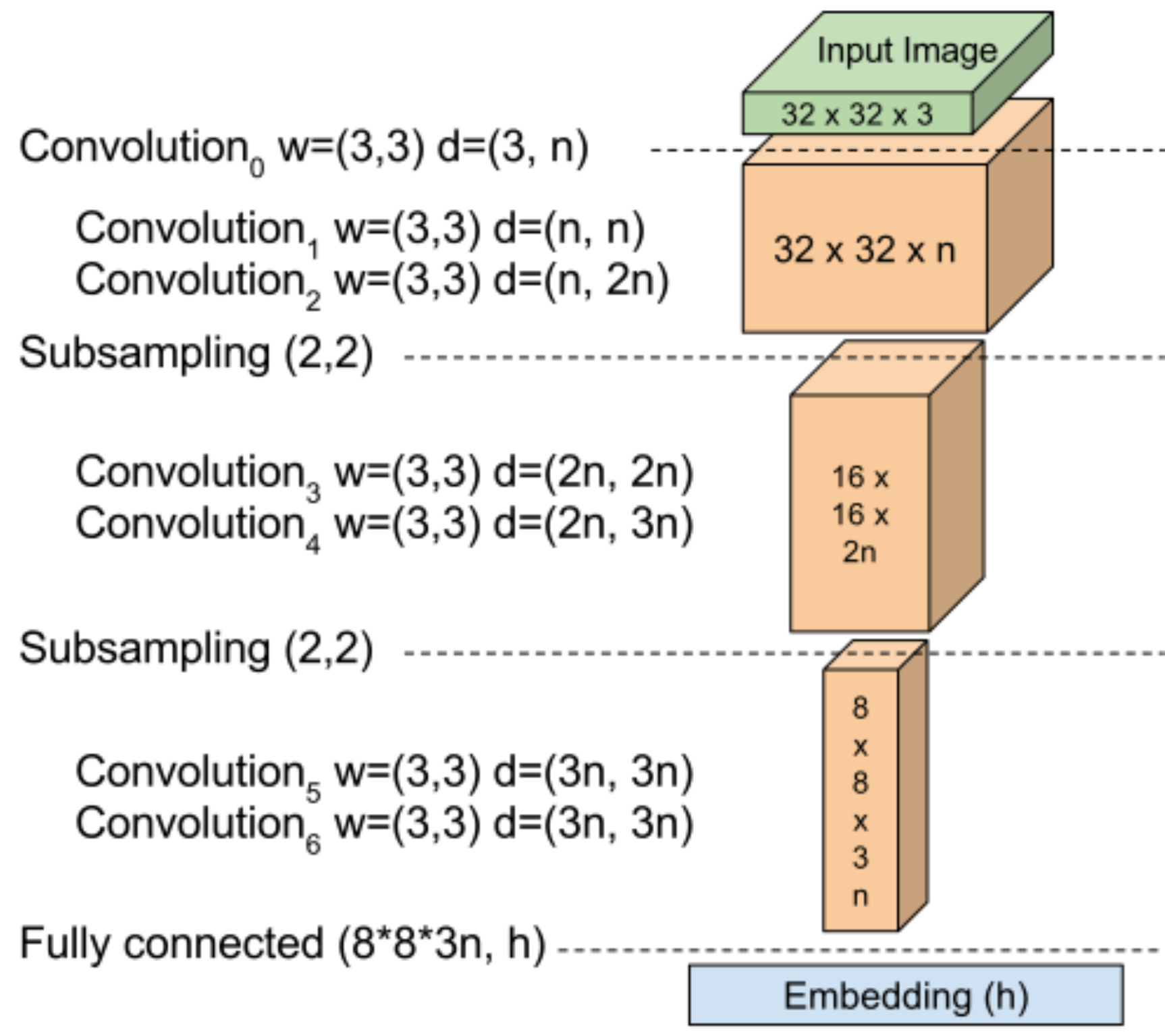}
\par\end{centering}

\begin{centering}

\par\end{centering}

}\caption{Network architecture for the generator and discriminator.\label{fig:Network}}

\end{figure}

The structure is shown in figure \ref{fig:Network}. We used $3\times3$
convolutions with exponential linear units \cite{clevert2015fast_elu}
(ELUs) applied at their outputs. Each layer is repeated a number of
times (typically 2). We observed that more repetitions led to even
better visual results. The convolution filters are increased linearly
with each down-sampling. Down-sampling is implemented as sub-sampling
with stride 2 and up-sampling is done by nearest neighbor. At the
boundary between the encoder and the decoder, the tensor of processed
data is mapped via fully connected layers, not followed by any non-linearities,
to and from an embedding state $h\in\mathbb{R}^{N_{h}}$ where $N_{h}$
is the dimension of the auto-encoder's hidden state. 

The generator $G:\mathbb{R}^{N_{z}}\mapsto\mathbb{R}^{N_{x}}$ uses
the same architecture (though not the same weights) as the discriminator
decoder. We made this choice only for simplicity. The input state
is $z\in\left[-1,1\right]^{N_{z}}$ sampled uniformly.

\subsubsection{Optional improvements}

This simple architecture achieves high quality results and demonstrates
the robustness of our technique.

Further, optional, refinements aid gradient propagation and produce
yet sharper images. Taking inspiration from deep residual networks
\cite{he2016deep}, we initialize the network using vanishing residuals:
for successive same sized layers, the layer's input is combined with
its output: $in_{x+1}=carry\times in_{x}+(1-carry)\times out_{x}$.
In our experiments, we start with $carry=1$ and progressively decrease
it to 0 over 16000 steps (one epoch). 

We also introduce skip connections \cite{he2016deep,srivastava2015highway,huang2016densely}
to help gradient propagation \cite{bengio1994learning}. The first
decoder tensor $h0$ is obtained from projecting $h$ to an $8\times8\times n$
tensor. After each upsampling step, the output is concatenated with
$h0$ upsampled to the same dimensions. This creates a skip connection
between the hidden state and each successive upsampling layer of the
decoder.

We did not explore other techniques typically used in GANs, such as
batch normalization, dropout, transpose convolutions or exponential
growth for convolution filters, though they might further improve
upon these results.

\section{Experiments}

\subsection{Setup}

We trained our model using Adam with an initial learning rate of $0.0001$,
decaying by a factor of $2$ when the measure of convergence stalls.
Modal collapses or visual artifacts were observed sporadically with
high initial learning rates, however simply reducing the learning
rate was sufficient to avoid them. We trained models for varied resolutions
from $32$ to $256$, adding or removing convolution layers to adjust
for the image size, keeping a constant final down-sampled image size
of $8\times8$. We used $N_{h}=N_{z}=64$ in most of our experiments
with this dataset.

Our biggest model for $128\times128$ images used a convolution with
$n=128$ filters and had a total of $17.3\times10^{6}$ trainable
parameters. Training time was about $2.5$ days on four P100 GPUs.
Smaller models of size $32\times32$ could train in a few hours on
a single GPU.

We use a dataset of $360K$ celebrity face images for training in
place of CelebA \cite{liu2015faceattributes_celeba}. This dataset
has a larger variety of facial poses, including rotations around the
camera axis. These are more varied and potentially more difficult
to model than the aligned faces from CelebA, presenting an interesting
challenge. We preferred the use of faces as a visual estimator since
humans excel at identifying flaws in faces.
\begin{figure}
\begin{centering}
\subfloat[EBGAN (64x64)\label{fig:EBGAN-(64x64)}]{\centering{}\includegraphics[width=6.8cm]{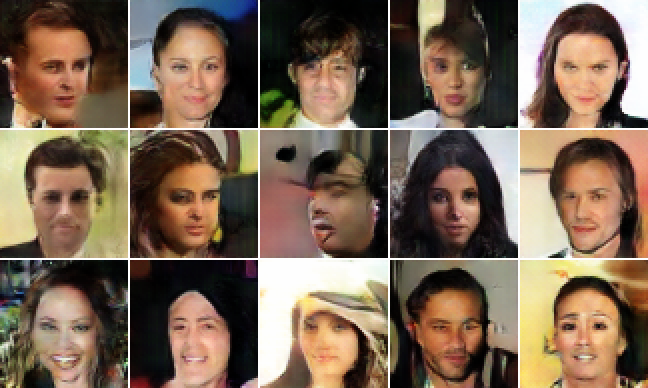}}\enskip{}\subfloat[Our results (128x128)\label{fig:Our-results-(128x128)} ]{\centering{}\includegraphics[width=6.8cm]{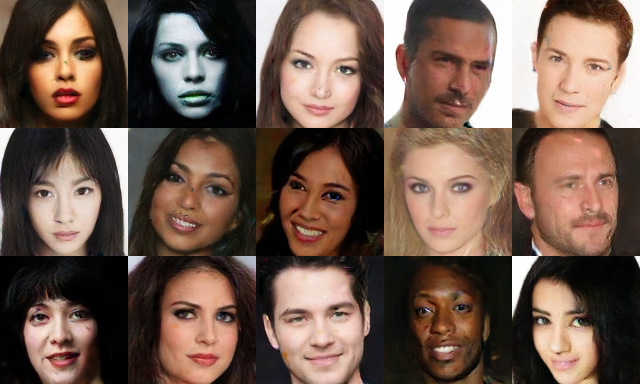}}
\par\end{centering}

\centering{}\caption{Random samples comparison \label{fig:rand}}
\end{figure}

\subsection{Image diversity and quality}

Figure \ref{fig:Our-results-(128x128)} shows some representative
samples drawn uniformly from $z$ at resolutions of $128\times128$.
Higher resolution images, while maintaining coherency, tend to lose
sharpness, but this may be improved upon with additional hyper-parameter
explorations. To our knowledge these are the first anatomically coherent
high-resolution results except for Stacked GANs \cite{zhang2016stackgan}
which has shown some promise for flowers and birds at up to $256\times256$. 

We observe varied poses, expressions, genders, skin colors, light
exposure, and facial hair. However we did not see glasses, we see
few older people and there are more women than men. For comparison
we also displayed some EBGAN \cite{zhao2016energy_ebgan} results
in figure \ref{fig:EBGAN-(64x64)}. We must keep in mind that these
are trained on different datasets so direct comparison is difficult.

\begin{figure}
\centering{}\includegraphics[bb=-100bp 0bp 2148bp 784bp,width=9cm]{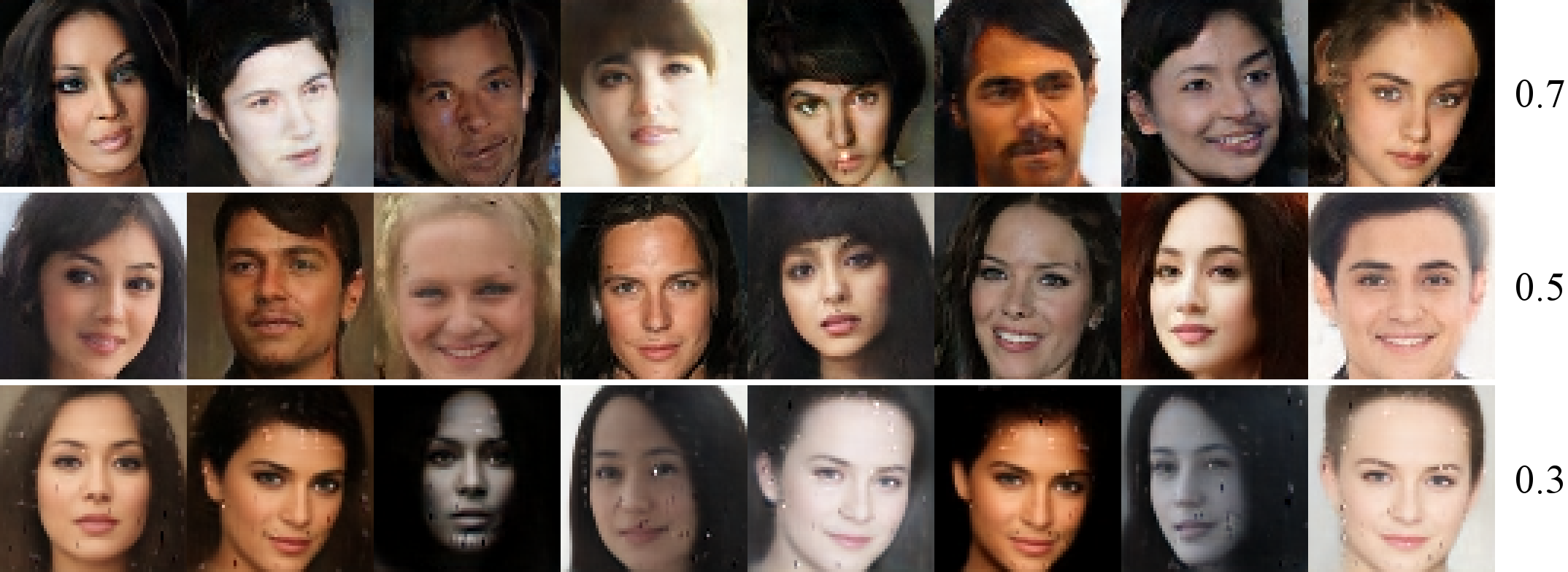}\caption{Random 64x64 samples at varying $\gamma\in\left\{ 0.3,0.5,0.7\right\} $\label{fig:randvar}}
\end{figure}

In Figure \ref{fig:randvar}, we compared the effect of varying $\gamma$.
The model appears well behaved, still maintaining a degree of image
diversity across the range of values. At low values, the faces look
overly uniform. Variety increases with $\gamma$ but so do artifacts.
Our observations seem to contradict those of \cite{poole2016improved}
that diversity and quality were independent.

\subsection{Space continuity}

To estimate the modal coverage of our generator we take real images
and find their corresponding $z_{r}$ embedding for the generator.
This is done using Adam to find a value for $z_{r}$ that minimizes
$e_{r}=|x_{r}-G(z_{r})|$. Mapping to real images is not the goal
of the model but it provides a way of testing its ability to generalize.
By interpolating the $z_{r}$ embeddings between two real images,
we verify that the model generalized the image contents rather than
simply memorizing them.

\begin{figure}
\begin{centering}
\subfloat[ALI \cite{dumoulin2016adversarially} (64x64)\label{fig:ALI-interpolation-(64x64)}]{\begin{centering}
\includegraphics[width=13.9cm]{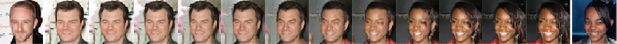}
\par\end{centering}

}
\par\end{centering}

\begin{centering}
\subfloat[Conditional PixelCNN \cite{oord2016conditional} (32x32)\label{fig:PixelCNN-interpolation-(64x64)}]{\begin{centering}
\includegraphics[width=13.9cm]{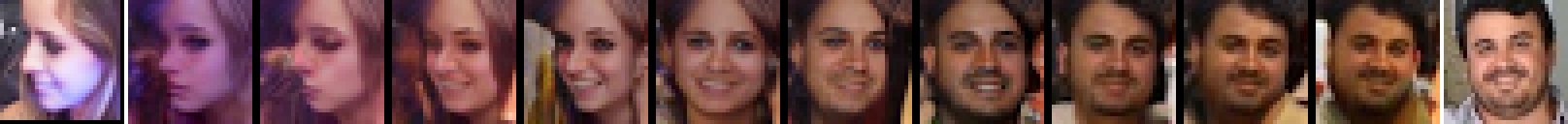}
\par\end{centering}

}
\par\end{centering}

\begin{centering}
\subfloat[Our results (128x128 with 128 filters)\label{fig:Our-results-(128x128}]{\begin{centering}
\includegraphics[width=13.9cm]{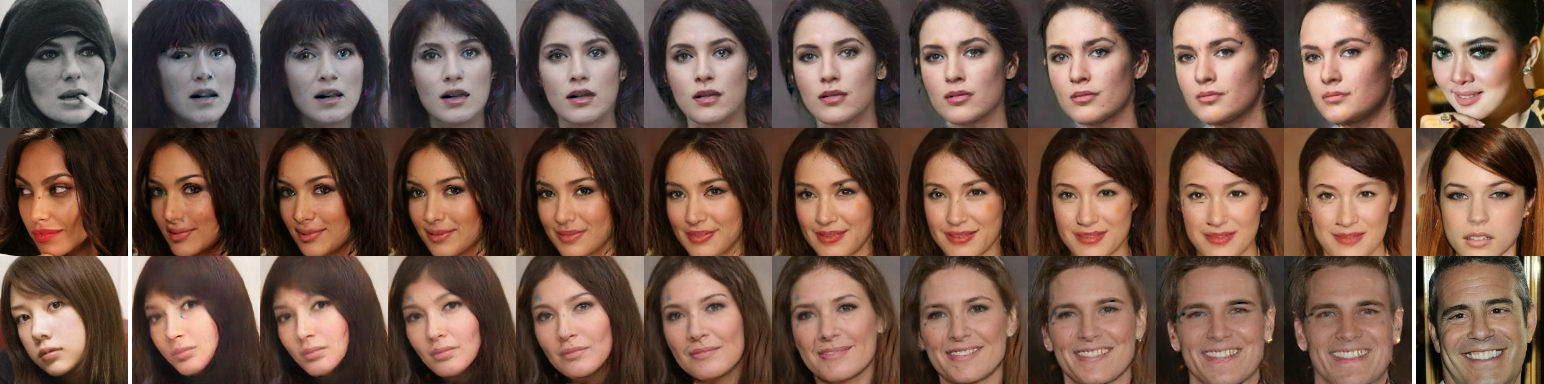}
\par\end{centering}

}
\par\end{centering}

\begin{centering}
\subfloat[Mirror interpolations (our results 128x128 with 128 filters)\label{fig:Mirror-interpolations-(our}]{\begin{centering}
\includegraphics[width=13.9cm]{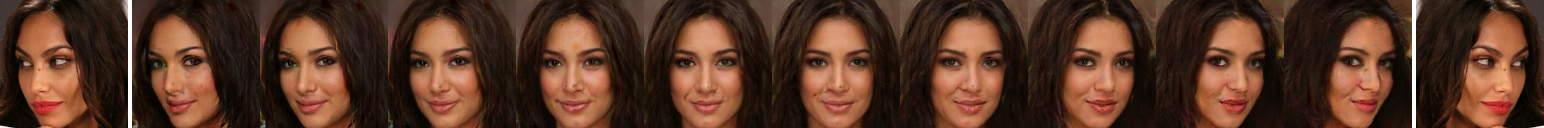}
\par\end{centering}

}
\par\end{centering}

\centering{}\caption{Interpolations of real images in latent space\label{fig:Interpolations}}
\end{figure}

Figure \ref{fig:Our-results-(128x128} displays interpolations on
$z_{r}$ between real images at $128\times128$ resolution; these
images were not part of the training data. The first and last columns
contain the real images to be represented and interpolated. The images
immediately next to them are their corresponding approximations while
the images in-between are the results of linear interpolation in $z_{r}$.
For comparison with the current state of the art for generative models,
we included ALI \cite{dumoulin2016adversarially} results at $64\times64$
(figure \ref{fig:ALI-interpolation-(64x64)}) and conditional PixelCNN
\cite{oord2016conditional} results at $32\times32$ (figure \ref{fig:PixelCNN-interpolation-(64x64)})
both trained on different data sets (higher resolutions were not available
to us for these models). In addition figure \ref{fig:Mirror-interpolations-(our}
showcases interpolation between an image and its mirror.

\begin{figure}[b]
\centering{}\includegraphics[bb=20bp 0bp 884bp 324bp,width=14cm]{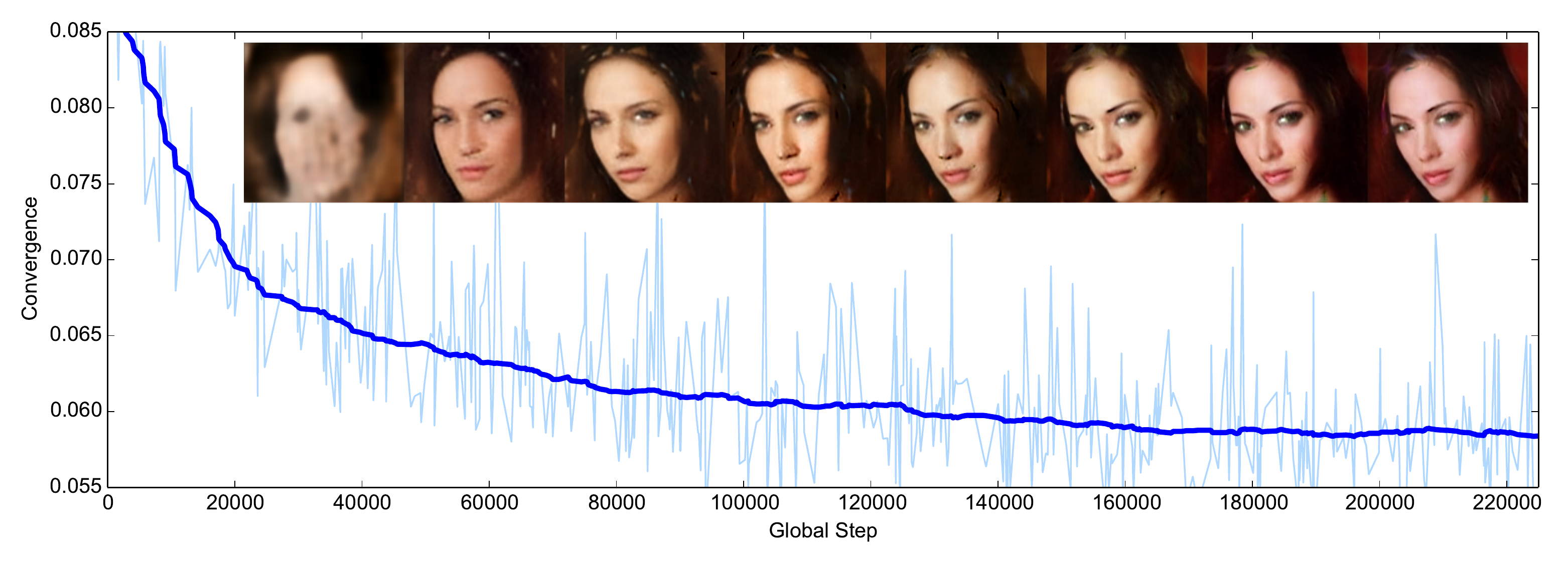}\caption{Quality of the results w.r.t. the measure of convergence (128x128
with 128 filters)\label{fig:convqual}}
\end{figure}
Sample diversity, while not perfect, is convincing; the generated
images look relatively close to the real ones. The interpolations
show good continuity. On the first row, the hair transitions in a
natural way and intermediate hairstyles are believable, showing good
generalization. It is also worth noting that some features are not
represented such as the cigarette in the left image. The second and
last rows show simple rotations. While the rotations are smooth, we
can see that profile pictures are not captured as well as camera facing
ones. We assume this is due to profiles being less common in our dataset.
Finally the mirror example demonstrates separation between identity
and rotation. A surprisingly realistic camera-facing image is derived
from a single profile image.

\subsection{Convergence measure and image quality}

The convergence measure $\mathcal{M}_{global}$ was conjectured earlier
to measure the convergence of the BEGAN model. As can be seen in figure
\ref{fig:convqual} this measure correlates well with image fidelity.
We can also see from this plot that the model converges quickly, just
as was originally reported for EBGANs. This seems to confirm the fast
convergence property comes from pixel-wise losses.
\begin{figure}
\begin{centering}
\subfloat[Starved generator ($z=16$ and $h=128$)\label{fig:Boosted-discriminator}]{\centering{}\includegraphics[bb=0bp 0bp 256bp 0bp,width=6.5cm]{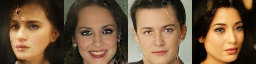}}\quad{}\subfloat[Starved discriminator ($z=128$ and $h=16$)\label{fig:Boosted-generator}]{\centering{}\includegraphics[width=6.5cm]{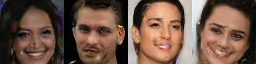}}
\par\end{centering}

\caption{Advantaging one network over the other\label{fig:Advantaging}}
\end{figure}

\subsection{Equilibrium for unbalanced networks}

To test the robustness of the equilibrium balancing technique, we
performed an experiment advantaging the discriminator over the generator,
and vice versa. Figure \ref{fig:Advantaging} displays the results.

By maintaining the equilibrium the model remained stable and converged
to meaningful results. The image quality suffered as expected with
low dimensionality of $h$ due to the reduced capacity of the discriminator.
Surprisingly, reducing the dimensionality of $z$ had relatively little
effect on image diversity or quality.

\subsection{Numerical experiments}

\begin{wraptable}[12]{O}{0.45\columnwidth}%
\noindent \centering{}%
\begin{tabular*}{5cm}{@{\extracolsep{\fill}}lr@{\extracolsep{0pt}.}l}
\toprule 
Method (unsupervised) & \multicolumn{2}{c}{Score}\tabularnewline
\midrule
\midrule 
Real data & 11&24\tabularnewline
\midrule
DFM \cite{warde2017improving} & 7&72\tabularnewline
\textbf{BEGAN (ours)} & \textbf{5}&\textbf{62}\tabularnewline
ALI \cite{dumoulin2016adversarially} & 5&34\tabularnewline
Improved GANs \cite{salimans2016improved} & 4&36\tabularnewline
MIX + WGAN \cite{arora2017generalization} & 4&04\tabularnewline
\bottomrule
\end{tabular*}\caption{Inception scores (higher is better)\label{tab:Inception-scores}}
\end{wraptable}%

To measure quality and diversity numerically, we computed the inception
score \cite{salimans2016improved} on CIFAR-10 images. The inception
score is a heuristic that has been used for GANs to measure single
sample quality and diversity on the inception model. We train an unconditional
version of our model and compare to previous unsupervised results.
The goal is to generate a distribution that is representative of the
original data.

A comparison to similar works on models trained entirely unsupervised
is shown in table \ref{tab:Inception-scores}. With the exception
of Denoising Feature Matching \cite{warde2017improving} (DFM), our
score is better than other GAN techniques that directly aim to match
the data distribution. This seems to confirm experimentally that matching
loss distributions of the auto-encoder is an effective indirect method
of matching data distributions. DFM appears compatible with our method
and combining them is a possible avenue for future work.

\section{Conclusion}

There are still many unexplored avenues. Does the discriminator have
to be an auto-encoder? Having pixel-level feedback seems to greatly
help convergence, however using an auto-encoder has its drawbacks:
what latent space size is best for a dataset? When should noise be
added to the input and how much? What impact would using other varieties
of auto-encoders such Variational Auto-Encoders\cite{kingma2013auto_vae}
(VAEs) have?

More fundamentally, we note that our objective bears a superficial
resemblance to the WGAN \cite{arjovsky2017wasserstein} objective.
Is the auto-encoder combined with the equilibrium concept fulfilling
a similar bounding functionality as the K-Lipschitz constraint in
the WGAN formulation?

We introduced BEGAN, a GAN that uses an auto-encoder as the discriminator.
Using proportional control theory, we proposed a novel equilibrium
method for balancing adversarial networks. We believe this method
has many potential applications such as dynamically weighing regularization
terms or other heterogeneous objectives. Using this equilibrium method,
the network converges to diverse and visually pleasing images. This
remains true at higher resolutions with trivial modifications. Training
is stable, fast and robust to parameter changes. It does not require
a complex alternating training procedure. Our approach provides at
least partial solutions to some outstanding GAN problems such as measuring
convergence, controlling distributional diversity and maintaining
the equilibrium between the discriminator and the generator. While
we could partially control the diversity of generator by influencing
the discriminator, there is clearly still room for improvement.

\section*{Acknowledgements}

We would like to thank Jay Han, Llion Jones and Ankur Parikh for their
help with the manuscript, Jakob Uszkoreit for his constant support,
Wenze Hu, Aaron Sarna and Florian Schroff for technical support. Special
thanks to Grant Reaber for his in-depth feedback on Wasserstein distance
computation.

\pagebreak{}\bibliographystyle{plain}
\bibliography{began}

\end{document}